\documentclass{article}
\usepackage[english]{babel}
\usepackage[letterpaper,top=2cm,bottom=2cm,left=3cm,right=3cm]{geometry}
\usepackage{amsmath, amssymb}
\usepackage{graphicx}
\usepackage[colorlinks=true, allcolors=blue]{hyperref}
\usepackage{algorithm}
\usepackage{algorithmic}
\usepackage{pifont}
\usepackage{soul}      
\usepackage{xcolor}    
\usepackage{natbib} 
\sethlcolor{yellow}
\usepackage{hyperref}

\title{TA-RNN-Medical-Hybrid: A Time-Aware and Interpretable Framework for Mortality Risk Prediction}
\author{
Zahra Jafari$^{1}$,
Azadeh Zamanifar*$^{2}$,
Amirfarhad Farhadi$^{1}$\\[6pt]
$^{1}$Department of Computer Engineering, SRC.,Islamic Azad university, Tehran, Iran\\
$^{2}$School of Computer Engineering, Iran university of Science and technology, Tehran, Iran\\
\texttt{azamanifar@iau.ac.ir}
}
\date{}

\begin{document}
\maketitle

\begin{abstract}
Accurate and interpretable mortality risk prediction in intensive care units (ICUs) remains a critical challenge due to the irregular temporal structure of electronic health records (EHRs), the complexity of longitudinal disease trajectories, and the lack of clinically grounded explanations in many data-driven models. To address these challenges, we propose \textit{TA-RNN-Medical-Hybrid}, a time-aware and knowledge-enriched deep learning framework that jointly models longitudinal clinical sequences and irregular temporal dynamics through explicit continuous-time encoding, along with standardized medical concept representations.
The proposed framework extends time-aware recurrent modeling by integrating explicit continuous-time embeddings that operate independently of visit indexing, SNOMED-based disease representations, and a hierarchical dual-level attention mechanism that captures both visit-level temporal importance and feature/concept-level clinical relevance. This design enables accurate mortality risk estimation while providing transparent and clinically meaningful explanations aligned with established medical knowledge.
We evaluate the proposed approach on the MIMIC-III critical care dataset and compare it against strong time-aware and sequential baselines. Experimental results demonstrate that TA-RNN-Medical-Hybrid consistently improves predictive performance in terms of AUC, accuracy, and recall-oriented F$_2$-score. Moreover, qualitative analysis shows that the model effectively decomposes mortality risk across time and clinical concepts, yielding interpretable insights into disease severity, chronicity, and temporal progression.
Overall, the proposed framework bridges the gap between predictive accuracy and clinical interpretability, offering a scalable and transparent solution for high-stakes ICU decision support systems.
\end{abstract}
\section{Introduction}
Mortality prediction in intensive care units (ICUs) has attracted sustained research attention due to its critical role in clinical decision support and resource management. The widespread availability of large-scale electronic health record (EHR) datasets, such as MIMIC-III, has facilitated the development of data-driven predictive models aimed at improving early risk stratification and patient outcomes \cite{johnson2016mimic}. Early studies predominantly relied on conventional machine learning techniques using handcrafted features extracted at admission or aggregated over short observation windows. Logistic regression, support vector machines, random forests, and gradient boosting models have been widely applied across diverse ICU cohorts, including sepsis patients, postoperative cases, pediatric respiratory diseases, acute pancreatitis, paralytic ileus, and cardiac arrest populations \cite{gao2024sepsis, rahman2024sepsis3, an2025sepsisckd, postopAI2025, respiratory2024ped, wei2025pancreatitis, bioengineering2024PI, li2025cardiacarrest}. Although these approaches achieved competitive performance compared to traditional clinical scoring systems, their reliance on static or early-aggregated representations limited their ability to capture complex temporal disease progression patterns inherent in ICU settings.

To overcome these limitations, subsequent research increasingly adopted deep learning models capable of learning nonlinear interactions among high-dimensional clinical variables. Neural network–based approaches, including multilayer perceptrons and recurrent neural networks (RNNs), demonstrated improved predictive accuracy across various ICU subpopulations such as COVID-19 patients, liver cancer cohorts, and surgical ICU admissions \cite{jmir2025covid, zheng2025livercancer, zhang2025surgery}. Nevertheless, many of these models discretize longitudinal EHR data into fixed time windows or process temporal sequences in a coarse-grained manner, thereby insufficiently reflecting the irregular sampling frequency and heterogeneous temporal structure of real-world ICU data.

Recognizing the importance of temporal dynamics, recent studies have proposed dynamic and real-time mortality prediction frameworks that explicitly leverage longitudinal EHR trajectories. Continuously updated risk estimation has consistently demonstrated superior performance compared to static models \cite{tbalsystems2025, jagesar2024comparative}. Among sequential modeling approaches, attention-enhanced recurrent architectures have shown particular promise. TA-RNN introduced an attention-based time-aware recurrent neural network that explicitly models irregular temporal gaps between clinical visits while assigning importance weights to sequential observations \cite{al_olaimat2024ta_rnn}. By incorporating time-aware decay mechanisms and visit-level attention, TA-RNN established a strong baseline for irregular EHR modeling. However, such models remain predominantly data-driven and provide limited integration of structured medical knowledge, constraining their ability to support clinically grounded reasoning.

Beyond recurrent architectures, transformer-based models have recently demonstrated strong performance in longitudinal EHR modeling. BEHRT \cite{li2020behrt} adapts the BERT architecture to structured EHR sequences by learning contextualized disease representations across visits. Med-BERT \cite{rasmy2021medbert} further extends this paradigm through large-scale self-supervised pretraining on structured medical codes, enabling transferable clinical representations across tasks. In parallel, GRU-D \cite{che2018grud} introduced a decay-based recurrent mechanism specifically designed to handle irregularly sampled time-series data in healthcare settings. While these approaches provide strong temporal representation learning capabilities, they either remain primarily data-driven without explicit integration of standardized medical ontologies or do not jointly address continuous-time irregularity and multi-level clinical interpretability. These limitations highlight the need for hybrid frameworks that integrate time-aware modeling with structured medical knowledge and hierarchical explanation mechanisms.

In parallel with advances in temporal modeling, interpretability and clinical transparency have emerged as central requirements for high-stakes ICU prediction systems. Recent studies emphasize interpretable machine learning frameworks that assess fairness, feature attribution, and clinical trustworthiness, particularly in disease-specific ICU populations such as hypertension and atrial fibrillation \cite{frontiers2025hypertension, chen2025AFhypertension}. Concept-driven and reasoning-inspired approaches further attempt to approximate clinician decision-making processes and improve explainability \cite{protodoctor2025}. Transformer-based architectures with per-case modality attribution have also been proposed to enhance transparency in early ICU mortality prediction \cite{clinicalTransf2025}. Despite these advances, most explanation mechanisms remain weakly aligned with standardized medical ontologies, limiting their semantic consistency and clinical interpretability.

Another emerging direction involves multimodal learning and heterogeneous data fusion. Integrating structured EHR variables with free-text clinical notes and additional modalities has been shown to improve robustness and predictive performance \cite{multimodalEHR2025}. However, such approaches often require complex fusion strategies and substantial computational resources, potentially limiting scalability and real-time deployment in clinical environments.

Recent systematic reviews confirm a clear trend toward attention-based, longitudinal, and interpretable models for ICU mortality prediction, while highlighting persistent gaps in modeling irregular temporal dynamics, incorporating domain-specific medical knowledge, and providing clinically meaningful explanations \cite{systematic2025}. These observations motivate the development of hybrid frameworks that jointly combine time-aware sequential modeling, knowledge-driven representations, and multi-level interpretability.

Modeling irregular longitudinal EHRs remains a fundamental challenge in ICU mortality prediction. Time-aware recurrent neural networks have emerged as an effective solution by explicitly accounting for non-uniform temporal intervals between clinical observations. Among these, TA-RNN demonstrated strong performance by jointly modeling temporal dynamics and visit-level importance. However, it does not explicitly support ontology-aligned disease reasoning or hierarchical interpretability.

In addition to prior sequential and transformer-based approaches, recent studies have investigated the application of Generative AI in clinical and healthcare contexts. These works demonstrate the potential of Generative AI to enhance workflow automation, support decision-making, and enable scalable healthcare solutions~\cite{taheri2025genai, rashidieranjbar2025genai}. Foundational research in related domains, such as Persian text summarization using fractal theory~\cite{ofighy2011fractal} and resilient connectivity restoration in wireless sensor and actor networks~\cite{zamanifar2009wsan}, also provides methodological insights relevant to intelligent data processing and system reliability. Collectively, these studies emphasize the importance of integrating domain knowledge, temporal reasoning, and interpretable modeling, motivating the development of the proposed hybrid framework.

Table~\ref{tab:comparison} summarizes the key methodological characteristics of TA-RNN and representative ICU mortality prediction models, highlighting differences in temporal modeling, irregular time handling, attention design, external medical knowledge integration, and interpretability level. This comparison illustrates that while TA-RNN effectively addresses irregular temporal dynamics through time-aware recurrence and visit-level attention, it lacks mechanisms for disease-level reasoning and knowledge-guided representation learning. These limitations motivate the architectural extensions introduced in the proposed framework.

\begin{table}[t]
\centering
\caption{Comparison of representative ICU mortality prediction models}
\label{tab:comparison}
\resizebox{\textwidth}{!}{
\begin{tabular}{lcccccc}
\hline
\textbf{Method} &
\textbf{Temporal Modeling} &
\textbf{Irregular Time Handling} &
\textbf{Attention Mechanism} &
\textbf{External Medical Knowledge} &
\textbf{Interpretability Level} &
\textbf{Clinical Applicability} \\
\hline
Traditional ML Models
& Limited
& No
& No
& No
& Feature-level
& Moderate \\

Deep Learning (MLP/CNN)
& Limited
& No
& No
& No
& Low
& Moderate \\

RNN / GRU
& Sequential
& Partial
& No
& No
& Low
& Moderate \\

Transformer-based Models
& Long-range
& Partial
& Yes
& No
& Visit/Modality-level
& Moderate \\

TA-RNN \cite{al_olaimat2024ta_rnn}
& Time-aware sequential
& Yes
& Visit-level
& No
& Visit-level
& High \\

GRU-D \cite{che2018grud}
& \ding{51}
& \ding{51}
& No
& \ding{55}
& Hidden-state \\

BEHRT \cite{li2020behrt}
& \ding{51}
& Partial
& Self-attention
& \ding{55}
& Token-level \\

Med-BERT \cite{rasmy2021medbert}
& \ding{51}
& Partial
& Self-attention
& \ding{55}
& Token-level \\

\textbf{Proposed TA-RNN-Medical-Hybrid}
& \textbf{Time-aware sequential}
& \textbf{Yes}
& \textbf{Dual-level (visit + disease)}
& \textbf{Yes (SNOMED-based)}
& \textbf{Visit + disease-level}
& \textbf{High} \\
\hline
\end{tabular}
}
\end{table}

To address the aforementioned limitations, we propose \textbf{TA-RNN-Medical-Hybrid}, a unified knowledge-aware temporal framework for ICU mortality prediction. Rather than treating temporal modeling, medical knowledge integration, and interpretability as independent design components, the proposed architecture jointly models continuous-time dynamics and ontology-aligned disease representations within a single end-to-end trainable system. The framework preserves the core time-aware recurrent mechanism of TA-RNN while introducing three substantive methodological advancements.

First, we incorporate \textbf{ontology-aligned disease embeddings derived from SNOMED CT}, enabling the model to encode structured semantic relationships among clinical concepts beyond statistical co-occurrence patterns. This integration promotes semantically consistent representations across heterogeneous patient trajectories and improves robustness in sparse or noisy EHR settings.

Second, we introduce a \textbf{hierarchical dual-level attention mechanism} that jointly captures visit-level temporal importance and disease-level clinical relevance. Unlike TA-RNN, which assigns attention weights solely at the visit level, the proposed model decomposes mortality risk into fine-grained disease contributions, thereby supporting clinically meaningful and quantitatively grounded interpretability.

Third, we enhance temporal expressiveness through \textbf{explicit continuous-time encoding}, enabling simultaneous modeling of short-term physiological fluctuations and long-term disease progression. This design more faithfully represents irregular sampling patterns and temporal heterogeneity commonly observed in ICU EHR data.

Collectively, these components elevate the architecture from a purely sequential predictor to a \textbf{knowledge-aware, temporally expressive, and clinically transparent modeling framework}, while maintaining scalability and practical feasibility for real-world ICU deployment.

\noindent\textbf{Contributions.} The main contributions of this work are summarized as follows:

\begin{itemize}

\item We propose a unified knowledge-aware temporal modeling framework for ICU mortality prediction that jointly integrates continuous-time encoding and ontology-aligned disease representations within a single end-to-end trainable architecture.

\item We introduce ontology-aligned disease embeddings derived from SNOMED CT to incorporate structured medical semantics into recurrent EHR modeling, enabling semantically consistent patient representations beyond purely data-driven co-occurrence learning.

\item We develop a hierarchical dual-level attention mechanism that decomposes mortality risk into both visit-level temporal importance and disease-level clinical contributions, thereby enhancing clinically meaningful interpretability.

\item We design an explicit continuous-time encoding strategy to more faithfully model irregular sampling intervals and heterogeneous temporal dynamics in ICU EHR data.

\item Extensive experiments on the MIMIC-III benchmark demonstrate consistent improvements in discrimination performance and recall compared to strong time-aware baselines, while providing enhanced disease-level interpretability.

\end{itemize}

\section{Materials and Methods}
\subsection{Data Source and Preprocessing}
\subsubsection{Data Source}
This study is conducted using the the MIMIC-III database \cite{johnson2016mimic} (Medical Information Mart for Intensive Care III), a large-scale, publicly available electronic health record (EHR) database comprising de-identified clinical data collected from intensive care unit (ICU) admissions at Beth Israel Deaconess Medical Center between 2001 and 2012. The dataset includes detailed demographic information, longitudinal clinical observations, diagnoses, procedures, and mortality outcomes.

In this work, we construct a longitudinal cohort by aggregating patient visit records indexed by a unique patient identifier (RID). Each visit contains a set of clinical features and a binary mortality indicator. Patients with insufficient longitudinal information are excluded to ensure valid temporal modeling.
Patient visits are first ordered chronologically for each individual. To ensure a consistent indexing scheme across patients, visit indices are aligned to a regular six-month grid for reference purposes only. 
Importantly, this alignment does not enforce regular temporal spacing in the underlying data.
Actual temporal irregularity between visits is explicitly preserved and modeled through elapsed-time encoding, as described later.
Specifically, for each patient $i$, visits are reindexed as:
\begin{equation}
\text{VISCODE}_{i,t} = 6 \times t, \quad t = 0, 1, \dots, T_i - 1,
\end{equation}
where $T_i$ denotes the total number of visits for patient $i$. This transformation enables uniform temporal modeling while preserving visit order.
To support sequence-to-sequence prediction, patients are filtered based on a minimum number of visits. Given a historical window length $t_s$ and a future prediction horizon $f_{ts}$, only patients satisfying:
\begin{equation}
T_i \geq t_s + f_{ts}
\end{equation}
are retained. This guarantees that each training sample contains sufficient past information and corresponding future outcomes.

The dataset is split at the patient level into training and testing subsets to prevent information leakage across individuals. All subsequent preprocessing steps are performed independently on each split.
Let $\mathbf{x}_{i,t} \in \mathbb{R}^{F}$ denote the clinical feature vector of patient $i$ at visit $t$, where $F$ is the number of longitudinal features. Mortality indicators are explicitly excluded from the input feature space and retained solely as prediction targets.

For patients with identical numbers of visits, longitudinal records are grouped and transformed into a flattened representation:
\begin{equation}
\mathbf{X}_i = \left[ \mathbf{x}_{i,0}, \mathbf{x}_{i,1}, \dots, \mathbf{x}_{i,T_i-1} \right],
\end{equation}
where missing feature values are imputed with zeros. The corresponding mortality sequence is represented as:
\begin{equation}
\mathbf{y}_i = \left[ y_{i,0}, y_{i,1}, \dots, y_{i,T_i-1} \right], \quad y_{i,t} \in \{0,1\}.
\end{equation}
To generate supervised learning samples, a sliding window strategy is employed. For each patient sequence, a fixed-length history window of $t_s$ visits is used to predict mortality outcomes over a future horizon of $f_{ts}$ visits.

Formally, each training sample is defined as:
\begin{equation}
\mathcal{S}_{i,k} = \left( 
\{\mathbf{x}_{i,k}, \dots, \mathbf{x}_{i,k+t_s-1}\}, 
\{y_{i,k+t_s}, \dots, y_{i,k+t_s+f_{ts}-1}\}
\right),
\end{equation}
where $k$ indexes the starting visit of the window. This approach increases the effective number of samples while strictly preserving temporal causality.
Static demographic features are incorporated as auxiliary inputs. Categorical demographic variables are encoded using one-hot encoding, while numerical variables are normalized via z-score normalization:
\begin{equation}
\tilde{x} = \frac{x - \mu_{\text{train}}}{\sigma_{\text{train}} + \epsilon},
\end{equation}
where $\mu_{\text{train}}$ and $\sigma_{\text{train}}$ are computed exclusively from the training cohort, and $\epsilon$ is a small constant for numerical stability. Each patient is associated with a fixed demographic vector aligned with all generated temporal samples.
To explicitly model true temporal irregularity within visit sequences,
an elapsed time feature is constructed based on the actual time difference between visits.
Given a historical window, the elapsed time at step $t$ is defined as:
\begin{equation}
e_t =
\begin{cases}
0, & t = 0, \\
\frac{\Delta t_t}{\max_{j} \Delta t_j}, & t = 1, 2, \dots, t_s - 1,
\end{cases}
\end{equation}
The normalization factor is computed globally across the training set to maintain consistent temporal scaling, while preserving relative temporal variation within each historical window.
This formulation preserves irregular sampling patterns across visits while ensuring numerical stability and comparability through normalization.

The elapsed time sequence $\mathbf{E} = \{e_0, \dots, e_{t_s-1}\}$ is provided as an additional input channel, enabling time-aware sequence modeling.
The final dataset is represented as:
\begin{equation}
\mathcal{D} = \{ (X_j, E_j, D_j, Y_j) \}_{j=1}^{M},
\end{equation}
where $X_j \in \mathbb{R}^{t_s \times F}$ denotes the longitudinal feature sequence, $E_j \in \mathbb{R}^{t_s \times 1}$ is the elapsed time sequence, $D_j$ is the static demographic vector, and $Y_j \in \mathbb{R}^{f_{ts} \times 1}$ is the future mortality label sequence.
Although visit indices are placed on a regular grid for modeling convenience, the proposed framework remains fully time-aware, 
as all temporal reasoning is driven by explicit continuous-time embeddings rather than discrete visit positions.

\subsubsection{Preparation of SNOMED Embedding Matrix for ICD Codes}
Given a set of ICD codes $\mathcal{I} = \{i_1, i_2, \dots, i_N\}$ from the MIMIC dataset, we map each ICD code $i \in \mathcal{I}$ to one or more SNOMED concepts $\mathcal{S}_i = \{s_1, s_2, \dots, s_{M_i}\}$ using the official SNOMED mapping files.  
The normalized ICD code is computed as:
\[
\text{ICD}_{\text{norm}} = \text{Normalize}(\text{ICD}) = \text{strip, uppercase, remove prefix and punctuation}
\]

The mapping function is defined as:
\[
f_{\text{map}}: \text{ICD}_{\text{norm}} \to \mathcal{S}_i
\]

For each SNOMED concept $s \in \mathcal{S} = \bigcup_i \mathcal{S}_i$, we extract its description text $d_s$. A pretrained clinical language model (BioClinicalBERT) is used to embed the text into a dense vector space:
\[
\mathbf{t}_s = \text{TextEmbed}(d_s) \in \mathbb{R}^{d_t}
\]
where $d_t$ denotes the embedding dimension (e.g., $d_t = 768$).

To capture structural knowledge, we construct a graph $G = (V, E)$ in which each node $v_s \in V$ corresponds to a SNOMED concept $s$, and edges $(v_s, v_{s'}) \in E$ represent hierarchical (IS-A) or associative relations.  
Edge weights $w_{ss'}$ are assigned according to relation type:
\[
w_{ss'} = 
\begin{cases}
1.0 & \text{hierarchical (IS-A)}\\
0.8 & \text{synonym or associative}\\
0.5 & \text{otherwise}
\end{cases}
\]
These weights encode relative relational importance rather than precise semantic distances, and the model was empirically observed to be robust to moderate variations in their values. The SNOMED graph is constructed independently of patient outcomes and does not incorporate mortality-related labels or outcome statistics, thereby avoiding information leakage.

Structural embeddings are generated using the GraphSAGE algorithm:
\[
\mathbf{g}_s = \text{GraphSAGE}(G, s) \in \mathbb{R}^{d_g},
\]
where $d_g$ denotes the graph embedding dimension. GraphSAGE is employed due to its inductive capability and scalability, making it suitable for large and evolving medical ontologies such as SNOMED CT without requiring full-graph retraining.

The final SNOMED embedding is obtained by concatenating textual and structural representations:
\[
\mathbf{m}_s = [\mathbf{t}_s; \mathbf{g}_s] \in \mathbb{R}^{d_t + d_g}.
\]
By combining graph-based and text-based embeddings, the model mitigates excessive structural smoothing while preserving concept-specific semantic information.

Due to differences in coding granularity and coverage between ICD and SNOMED, a small fraction of ICD codes cannot be reliably mapped to any SNOMED concept. Let $\mathbf{E} \in \mathbb{R}^{(N+1) \times (d_t + d_g)}$ denote the resulting embedding matrix, where row $0$ is reserved for padding. For each ICD code $i$ with numeric identifier $n_i$, the embedding is computed as:
\[
\mathbf{E}[n_i + 1] =
\begin{cases}
\frac{1}{|\mathcal{S}_i|} \sum_{s \in \mathcal{S}_i} \mathbf{m}_s & \text{if } |\mathcal{S}_i| > 0,\\
\mathbf{e}_{\mathrm{unk}} & \text{otherwise}.
\end{cases}
\]

Unmapped ICD codes, accounting for less than 3\% of all diagnosis records, are assigned a shared learnable embedding $\mathbf{e}_{\mathrm{unk}}$ rather than random initialization. This shared representation reflects the absence of reliable semantic differentiation while preventing the introduction of uncontrolled noise during model training.

The embedding matrix $\mathbf{E}$ is saved as a NumPy array for direct use in downstream deep learning models.

\begin{figure}[t!]
    \centering
    \includegraphics[width=0.95\linewidth]{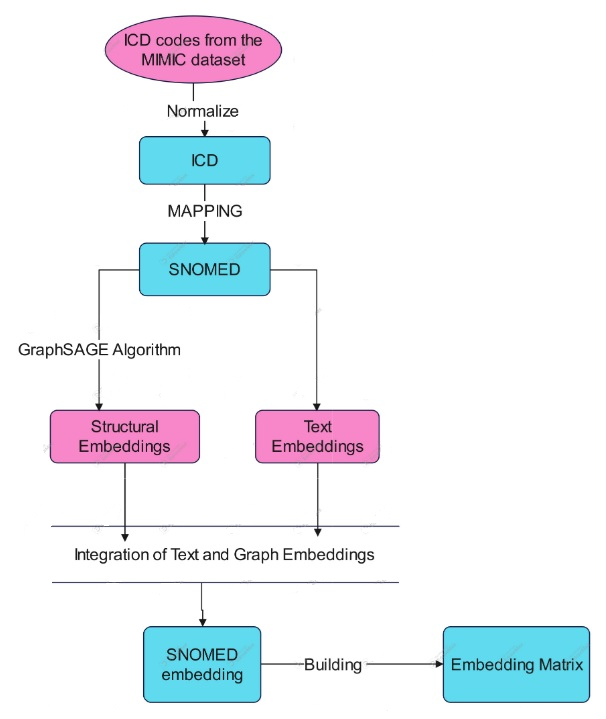}
    \caption{Pipeline for constructing the ICD-to-SNOMED embedding matrix. ICD codes are normalized and mapped to SNOMED concepts, followed by semantic text embedding using BioClinicalBERT and structural embedding via SNOMED relational graphs.}
    \label{fig:snomed_embedding_pipeline}
\end{figure}

Figure~\ref{fig:snomed_embedding_pipeline} illustrates the overall pipeline for constructing the ICD-to-SNOMED embedding matrix. The key steps are summarized as follows:
\begin{itemize}
    \item ICD codes are normalized and mapped to corresponding SNOMED concepts.
    \item Text embeddings capture semantic information from SNOMED concept descriptions.
    \item Graph embeddings encode structural and hierarchical relations among SNOMED concepts.
    \item Textual and structural embeddings are concatenated to form enriched SNOMED representations.
    \item The final embedding matrix enables downstream models to leverage both semantic and relational medical knowledge.
\end{itemize}

\subsection{Proposed Model}
\begin{figure}[!t]
    \centering
    \includegraphics[width=0.95\linewidth]{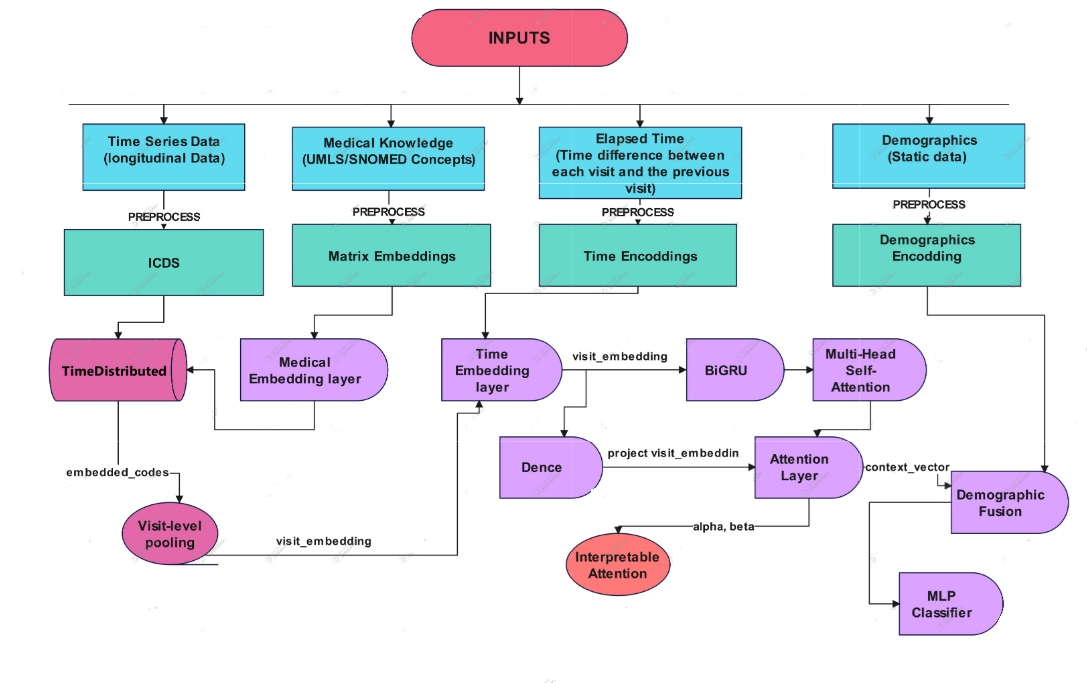}
    \caption{Overall architecture of the proposed TA-RNN-Medical-Hybrid framework.
    Diagnosis codes are first mapped to SNOMED-based embeddings and aggregated
    at the visit level. Explicit temporal embeddings are incorporated to handle
    irregular visit intervals. The resulting sequence is processed by a
    BiGRU and multi-head self-attention module, followed by a dual-level attention mechanism (visit + disease-level) and demographic fusion...
 to generate the final
    mortality risk prediction.}
    \label{fig:framework_overview}
\end{figure}

Figure~\ref{fig:framework_overview} provides an overview of the proposed
TA-RNN-Medical-Hybrid architecture, illustrating the integration of
knowledge-driven medical embeddings, explicit temporal encoding,
sequential modeling, and interpretable dual-level attention for mortality
risk prediction.

\subsubsection{Overview}

We propose a time-aware and interpretable deep learning framework, termed
\textit{TA-RNN-Medical-Hybrid}, for patient mortality risk prediction from longitudinal electronic health records (EHRs).
The model is designed to jointly capture temporal dynamics, sequential dependencies, and clinically meaningful feature importance.
It integrates four key components: knowledge-driven medical embeddings, explicit temporal encoding, sequential visit modeling, and an interpretable dual-level attention mechanism.
The overall computational procedure of the proposed TA-RNN-Medical-Hybrid framework,
including medical embedding, temporal encoding, sequential modeling, attention-based
interpretation, and mortality prediction, is summarized in the following pseudocode.
As illustrated in Figure~\ref{fig:framework_overview}, the proposed framework
jointly models longitudinal clinical trajectories, temporal irregularity,
and medical concept semantics while maintaining interpretability at both
the visit and disease levels.

\begin{algorithm}[t]
\caption{TA-RNN-Medical-Hybrid Framework for Mortality Risk Prediction}
\label{alg:ta_rnn}
\begin{algorithmic}[1]

\REQUIRE Longitudinal patient visits $\{c_{t,k}\}$, elapsed times $\{e_t\}$, demographic vector $\mathbf{d}$
\ENSURE Mortality risk prediction $\hat{y}$, visit attention $\{\alpha_t\}$, disease importance scores

\STATE Load pre-trained ICD embedding matrix $\mathbf{E}$ (frozen)

\FOR{$t = 1$ to $T$}
    \FOR{each diagnosis code $c_{t,k}$ in visit $t$}
        \STATE $\mathbf{z}_{t,k} \leftarrow \mathbf{E}[c_{t,k}]$
    \ENDFOR
    \STATE $\mathbf{v}_t \leftarrow \text{MeanPooling}(\{\mathbf{z}_{t,k}\})$
    \STATE $\mathbf{v}_t \leftarrow \mathbf{v}_t + \mathrm{TE}(e_t)$
\ENDFOR

\STATE $\{\mathbf{h}_1,\dots,\mathbf{h}_T\} \leftarrow \text{BiGRU}(\{\mathbf{v}_t\})$
\STATE $\mathbf{H}_{att} \leftarrow \text{MultiHeadSelfAttention}(\{\mathbf{h}_t\})$

\FOR{$t = 1$ to $T$}
    \STATE $\alpha_t \leftarrow \text{Softmax}(\mathbf{w}_\alpha^\top \mathbf{h}_t)$
    \STATE $\boldsymbol{\beta}_t \leftarrow \text{Softmax}(\tanh(\mathbf{W}_\beta \mathbf{h}_t))$
\ENDFOR

\STATE $\mathbf{c} \leftarrow \sum_{t=1}^{T} \alpha_t \cdot (\boldsymbol{\beta}_t \odot \mathbf{v}_t)$
\STATE $\mathbf{u} \leftarrow [\mathbf{c} \| \mathbf{d}]$
\STATE $\hat{y} \leftarrow \sigma(\text{MLP}(\mathbf{u}))$

\RETURN $\hat{y}, \{\alpha_t\}, \{\boldsymbol{\beta}_t\}$

\end{algorithmic}
\end{algorithm}

\subsubsection{Input Representation}

Each patient is represented by three input modalities:
(i) a sequence of diagnosis codes per visit,
(ii) a corresponding elapsed time sequence capturing irregular visit intervals,
and (iii) a vector of static demographic features.
Let a patient record consist of $T$ visits, each containing up to $K$ diagnosis codes.

\subsubsection{Knowledge-Driven Medical Embedding}

Diagnosis codes are mapped to fixed medical embeddings derived from external clinical knowledge.
Let $c_{t,k}$ denote the $k$-th diagnosis code in visit $t$.
Each code is embedded into a dense vector space using a pre-trained embedding matrix:
\[
\mathbf{z}_{t,k} = \mathrm{Embed}(c_{t,k}) \in \mathbb{R}^{d}.
\]

The embedding layer is kept frozen during training to preserve the encoded medical semantics.
To obtain a visit-level representation, diagnosis embeddings within each visit are aggregated via mean pooling:
\[
\mathbf{v}_t = \frac{1}{K} \sum_{k=1}^{K} \mathbf{z}_{t,k}.
\]

\subsubsection{Explicit Temporal Embedding}

To model irregular time intervals between visits independently of visit indexing, an explicit continuous-time embedding is incorporated.

Given the normalized elapsed time $e_t$ associated with visit $t$, a sinusoidal encoding is constructed as:
\[
\mathrm{TE}(e_t, i) =
\begin{cases}
\sin\left(\frac{e_t}{\lambda^{2i/d}}\right), & i \text{ even} \\
\cos\left(\frac{e_t}{\lambda^{2i/d}}\right), & i \text{ odd}
\end{cases}
\]
where $d$ denotes the embedding dimension and $\lambda$ is a scaling constant.

The temporal embedding is added to the visit representation:
\[
\tilde{\mathbf{v}}_t = \mathbf{v}_t + \mathrm{TE}(e_t).
\]

\subsubsection{Sequential Visit Encoding}

Temporal visit representations are processed using a two-layer bidirectional gated recurrent unit (BiGRU) network.
For each visit $t$, the BiGRU computes forward and backward hidden states:
\[
\overrightarrow{\mathbf{h}}_t = \mathrm{GRU}_f(\tilde{\mathbf{v}}_t), \quad
\overleftarrow{\mathbf{h}}_t = \mathrm{GRU}_b(\tilde{\mathbf{v}}_t).
\]

The visit-level hidden representation is obtained by concatenation:
\[
\mathbf{h}_t = [\overrightarrow{\mathbf{h}}_t \| \overleftarrow{\mathbf{h}}_t].
\]

Dropout regularization is applied between recurrent layers to mitigate overfitting.

\subsubsection{Multi-Head Self-Attention}

To capture global dependencies across visits, a multi-head self-attention (MHA) layer is applied to the sequence of BiGRU outputs:
\[
\mathbf{H}' = \mathrm{MHA}(\mathbf{H}, \mathbf{H}, \mathbf{H}),
\]
where $\mathbf{H} = \{\mathbf{h}_1, \dots, \mathbf{h}_T\}$.

A residual connection followed by layer normalization is employed:
\[
\mathbf{H}_{\text{att}} = \mathrm{LayerNorm}(\mathbf{H} + \mathbf{H}').
\]
The BiGRU is responsible for capturing local temporal dynamics and short-term physiological variations across adjacent visits.
In contrast, the self-attention mechanism explicitly models global temporal dependencies and long-range interactions over the entire visit sequence,
thereby enabling the model to jointly reason about short-term fluctuations and long-term disease progression.

\subsubsection{Interpretable Dual-Level Attention}

To enhance interpretability, the model employs a dual-level attention mechanism consisting of visit-level attention ($\alpha$) and feature-level attention ($\beta$).

The visit-level attention weights are computed as:
\[
\alpha_t = \frac{\exp(\mathbf{w}_\alpha^\top \mathbf{h}_t + b_\alpha)}
{\sum_{j=1}^{T} \exp(\mathbf{w}_\alpha^\top \mathbf{h}_j + b_\alpha)}.
\]

Feature-level attention is defined as:
\[
\boldsymbol{\beta}_t = \mathrm{softmax}\left(\tanh(\mathbf{W}_\beta \mathbf{h}_t + \mathbf{b}_\beta)\right).
\]

The final context vector is obtained by combining attention weights with projected visit embeddings:
\[
\mathbf{c} = \sum_{t=1}^{T} \alpha_t \left( \boldsymbol{\beta}_t \odot \mathbf{v}_t \right),
\]
where $\odot$ denotes element-wise multiplication.

\subsubsection{Demographic Fusion and Output Layer}

The context vector is concatenated with demographic features:
\[
\mathbf{u} = [\mathbf{c} \| \mathbf{d}],
\]
where $\mathbf{d}$ denotes the demographic feature vector.

The fused representation is passed through a multilayer perceptron (MLP) followed by a sigmoid activation to produce the final mortality risk prediction:
\[
\hat{y} = \sigma(\mathbf{W}_o \mathbf{u} + b_o).
\]

This design enables accurate mortality prediction while providing transparent and clinically meaningful interpretations at both the visit and disease levels.
\subsubsection{Parameter Learning and Evaluation Metrics}
\label{sec:training_evaluation}

All trainable parameters of the proposed TA-RNN-Medical-Hybrid framework were jointly optimized in an end-to-end manner using a customized weighted binary cross-entropy loss function. This loss formulation explicitly addresses the class imbalance inherent in ICU mortality prediction by assigning higher importance to correctly identifying positive (mortality) cases, thereby reducing false negatives and improving model sensitivity.

Formally, the loss function is defined as:
\begin{equation}
\mathcal{L} = -\frac{1}{N} \sum_{i=1}^{N}
\left[
\delta \, y_i \log(\hat{y}_i) +
(1 - \delta)(1 - y_i)\log(1 - \hat{y}_i)
\right],
\end{equation}
where $y_i \in \{0,1\}$ denotes the ground-truth mortality label for the $i$-th sample, $\hat{y}_i$ represents the predicted mortality probability, and $\delta \in (0,1)$ is a hyperparameter controlling the relative importance of positive class prediction. In all experiments, $\delta$ was treated as a tunable hyperparameter and set to $0.7$, emphasizing recall and sensitivity in high-risk ICU scenarios.

Model optimization was performed using the Adaptive Moment Estimation (Adam) optimizer with a learning rate of $0.001$. Early stopping was employed based on validation loss with a patience of 10 epochs to prevent overfitting, and the model parameters corresponding to the best validation performance were retained. Dropout regularization and $\ell_2$ weight decay were applied to recurrent and fully connected layers to further enhance generalization.

Key architectural and training hyperparameters—including batch size, number of training epochs, dropout rate, $\ell_2$ regularization coefficient, recurrent hidden size, RNN cell type, and the loss weighting parameter $\delta$—were selected through empirical tuning using a held-out validation subset comprising 20\% of the training data.

For performance evaluation, model predictions were assessed using accuracy, area under the receiver operating characteristic curve (AUC), and the $F_2$-score. The $F_2$-score was chosen as a primary metric due to its increased emphasis on recall, which is critical in ICU mortality prediction where failing to identify high-risk patients carries substantial clinical consequences. The $F_2$-score is defined as:
\begin{equation}
F_\beta = \frac{(1 + \beta^2) \cdot \text{Precision} \cdot \text{Recall}}
{\beta^2 \cdot \text{Precision} + \text{Recall}},
\end{equation}
where $\beta = 2$ in this study, assigning four times greater importance to recall compared to precision.

To ensure fair and clinically meaningful evaluation, the decision threshold for binary classification was optimized on validation data to maximize the $F_2$-score, rather than being fixed at a default probability of 0.5. All reported test results were computed using this optimized threshold. The combination of weighted loss optimization and recall-focused evaluation provides a robust assessment of model performance under clinically realistic ICU risk stratification requirements.

\subsubsection{Computational Complexity Analysis}

Let $T$ denote the number of visits per patient, $K$ the number of diagnosis codes per visit,
$d$ the embedding dimension, $h$ the hidden dimension of the recurrent units, $H$ the number
of attention heads, and $L$ the number of BiGRU layers. Note that temporal encoding operates on elapsed-time values and does not assume uniform sampling across visits.

The computational cost of ICD embedding and visit-level aggregation is
$\mathcal{O}(T K d)$.
The sinusoidal temporal encoding introduces a linear overhead of $\mathcal{O}(T d)$.

The dominant cost arises from the two-layer bidirectional GRU, which requires
$\mathcal{O}(L T h(d + h))$ operations.
The multi-head self-attention module incurs a cost of
$\mathcal{O}(H T^2 h)$ due to pairwise visit interactions.

The dual-level attention and final MLP layers introduce comparatively negligible overhead,
bounded by $\mathcal{O}(T h d)$.
Overall, the total computational complexity of the proposed model is given by:

\[
\mathcal{O}\Big(
T K d
+ L T h(d+h)
+ H T^2 h
\Big)
\]

In typical EHR settings where the number of visits per patient is moderate,
the model remains computationally efficient and scalable.

\section{Results and Discussion}

This section reports and interprets the experimental results of the proposed TA-RNN-Medical-Hybrid model in the context of ICU mortality prediction. The analysis emphasizes not only predictive performance but also the model’s ability to provide clinically meaningful explanations by capturing temporal dynamics and disease-level contributions. Through quantitative evaluation and qualitative interpretation, we demonstrate how the proposed framework supports transparent and actionable clinical decision-making.

\subsection{Risk Prediction and Model Interpretability}

In this subsection, we jointly examine the predictive accuracy and interpretability characteristics of the proposed framework. We first assess mortality risk prediction using standard evaluation metrics, and then analyze the model’s explanation mechanisms, including temporal attention and disease-level attribution. This combined analysis allows us to evaluate how predictive outcomes align with clinically plausible reasoning patterns.

\subsubsection{Mortality Risk Prediction and Clinical Stratification}

For each patient trajectory, the proposed framework outputs a continuous mortality risk score \( r \in [0,1] \), reflecting the model’s confidence in adverse outcomes. To facilitate clinical interpretability and downstream decision-making, risk scores are stratified into four clinically meaningful categories: low, moderate, high, and critical risk.

\[
\text{RiskCategory}(r) =
\begin{cases}
\text{Low}, & r < 0.20 \\
\text{Moderate}, & 0.20 \le r < 0.40 \\
\text{High}, & 0.40 \le r < 0.70 \\
\text{Critical}, & r \ge 0.70
\end{cases}
\]

\begin{figure}[t]
    \centering
    \includegraphics[width=0.9\linewidth]{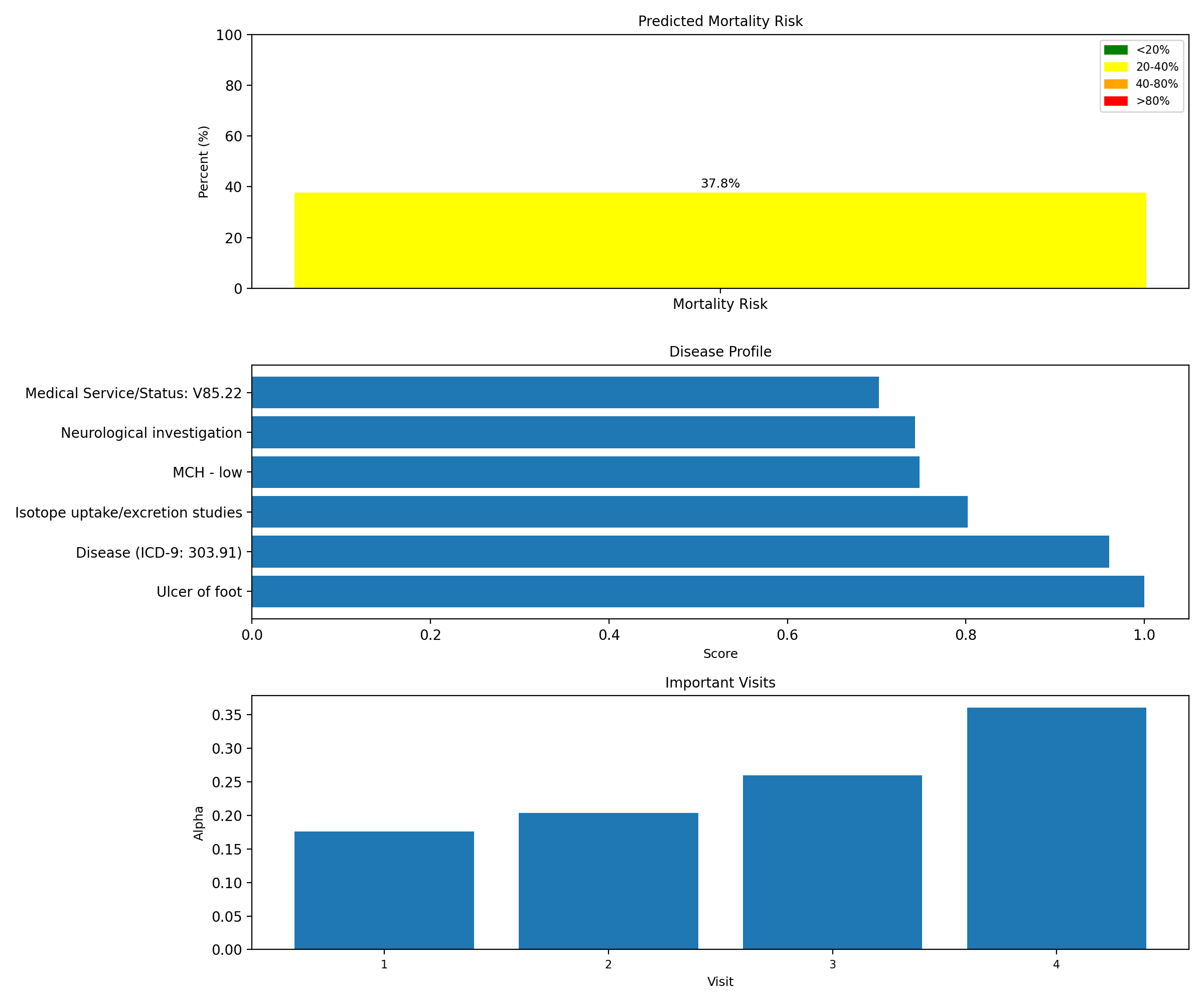}
    \caption{Summary of model interpretation outputs.
    The figure illustrates mortality risk stratification, visit-level attention distribution,
    and ranked disease-level contributions for representative ICU patients.}
    \label{fig:interpretation_summary}
\end{figure}

This stratification aligns with common ICU triage practices and allows risk estimates to be directly translated into actionable care pathways. As illustrated in Figure~\ref{fig:interpretation_summary}, the predicted risk distribution demonstrates clear separation across patient groups, indicating stable calibration and clinically meaningful discrimination.

\subsubsection{Temporal Contribution Analysis via Visit-Level Attention}

To understand how longitudinal information contributes to mortality prediction, we analyze visit-level attention coefficients \( \alpha_t \), which quantify the relative importance of each visit in the patient trajectory:

\[
\alpha_t \in [0,1], \quad \sum_{t=1}^{T} \alpha_t = 1.
\]

Higher attention values correspond to visits exerting greater influence on the final prediction. Figure~\ref{fig:interpretation_summary} visualizes attention weights across time, revealing that the model consistently emphasizes clinically critical periods, such as visits preceding physiological deterioration or major diagnostic changes.

Importantly, unlike uniform temporal aggregation, the learned attention patterns demonstrate strong temporal selectivity, indicating that TA-RNN-Medical-Hybrid effectively filters noisy or less informative visits while amplifying clinically salient events. This behavior is particularly valuable in ICU settings characterized by irregular sampling and heterogeneous visit density.

\subsubsection{Disease-Level Attribution through Knowledge-Enriched Embeddings}

A key contribution of the proposed framework lies in its ability to attribute mortality risk to individual SNOMED-based features within visits, in a clinically grounded manner. For a disease \( d \) appearing in visit \( t \), its contribution is defined as:

\[
C_d(t) = \alpha_t \cdot \left\lVert \mathbf{e}_d \right\rVert_2,
\]

where \( \mathbf{e}_d \) denotes the fixed SNOMED-based embedding of disease \( d \).
Aggregating across visits yields the cumulative disease importance score:

\[
S_d = \sum_{t=1}^{T} C_d(t),
\quad
\hat{S}_d = \frac{S_d}{\max_k S_k}.
\]

Figure~\ref{fig:interpretation_summary} presents ranked disease profiles for representative patients. Notably, highly ranked conditions often correspond to clinically recognized drivers of ICU mortality, supporting the semantic validity of the proposed knowledge-driven attribution mechanism. Compared to purely data-driven attention models, the integration of SNOMED embeddings enhances consistency and reduces spurious associations.

\subsubsection{Disease Severity and Chronicity Characterization}

Normalized feature/concept scores enable categorization into severity levels:

\[
\text{Severity}(\hat{S}_d) =
\begin{cases}
\text{Severe}, & \hat{S}_d > 0.70 \\
\text{Moderate}, & 0.40 < \hat{S}_d \le 0.70 \\
\text{Mild}, & \hat{S}_d \le 0.40.
\end{cases}
\]

This stratification reflects \emph{model-attributed importance} rather than conventional clinical staging, providing a complementary perspective on disease impact. Additionally, diseases appearing across multiple visits are identified as chronic, allowing differentiation between persistent conditions and acute episodic events.

The joint analysis of severity and chronicity reveals that chronic high-severity diseases disproportionately contribute to elevated mortality risk, while isolated acute diagnoses tend to exert transient influence. This distinction is particularly relevant for longitudinal care planning and resource allocation.

\subsubsection{Temporal Disease Progression Patterns}

To further examine feature dynamics, we analyze temporal contribution trajectories:

\[
P_d = [C_d(1), C_d(2), \dots, C_d(T)].
\] 

Linear trend estimation allows classification into increasing, decreasing, or stable progression patterns. Diseases with increasing trends frequently coincide with rising mortality risk, whereas decreasing trends often reflect treatment response or stabilization.

\begin{figure}[t]
    \centering
    \includegraphics[width=0.9\linewidth]{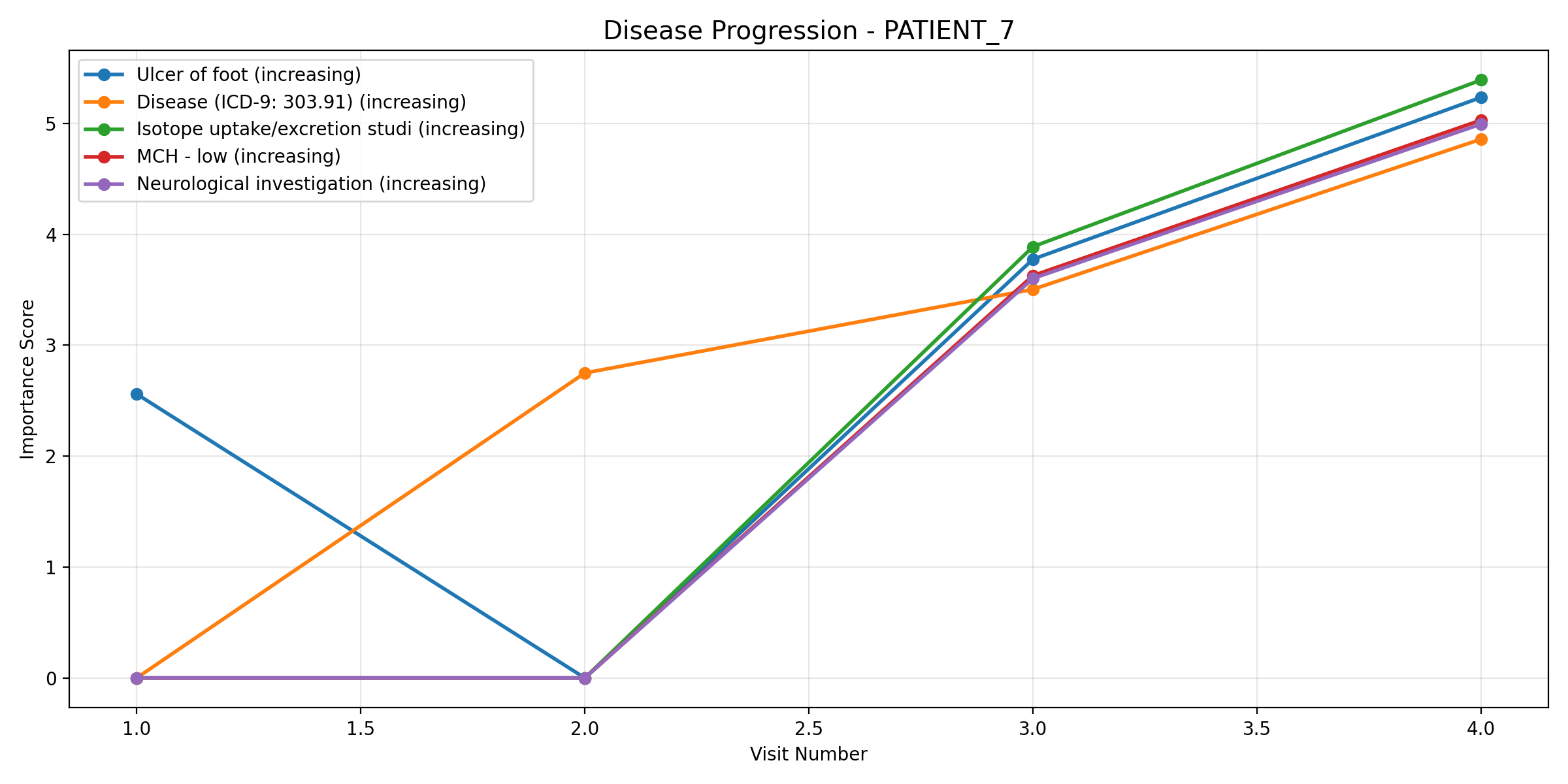}
    \caption{Temporal disease progression patterns derived from disease-level attribution scores.
    Increasing, decreasing, and stable contribution trends illustrate how individual diseases
    influence mortality risk over time.}
    \label{fig:disease_progression}
\end{figure}

Figure~\ref{fig:disease_progression} highlights representative progression patterns, demonstrating the framework’s ability to capture worsening, resolving, and stable disease trajectories. Identifying the peak contribution time \( t_{\text{peak}} \) further enables temporal localization of critical disease phases.

\subsubsection{Clinical Insight Generation and Decision Support Implications}

By integrating mortality risk stratification with disease-level attribution, the proposed framework supports automated generation of clinically actionable insights. For example, patients categorized as high or critical risk with multiple severe disease drivers may trigger escalation alerts, while moderate-risk patients with stable disease trajectories may warrant targeted monitoring.

Unlike black-box predictors, TA-RNN-Medical-Hybrid offers transparent explanations that align with clinical reasoning, potentially improving clinician trust and adoption. Importantly, explanations are grounded in standardized medical knowledge rather than purely statistical correlations.

\subsection{Discussion and Implications}

Overall, the results demonstrate that TA-RNN-Medical-Hybrid not only achieves reliable mortality risk estimation but also delivers clinically grounded, multi-level interpretability. In contrast to baseline time-aware models such as TA-RNN, which primarily provide visit-level importance scores, the proposed framework enables disease-level reasoning explicitly anchored in standardized medical knowledge, thereby narrowing the gap between predictive accuracy and clinical usability.
\begin{figure}[t!]
    \centering
    \includegraphics[width=0.95\linewidth]{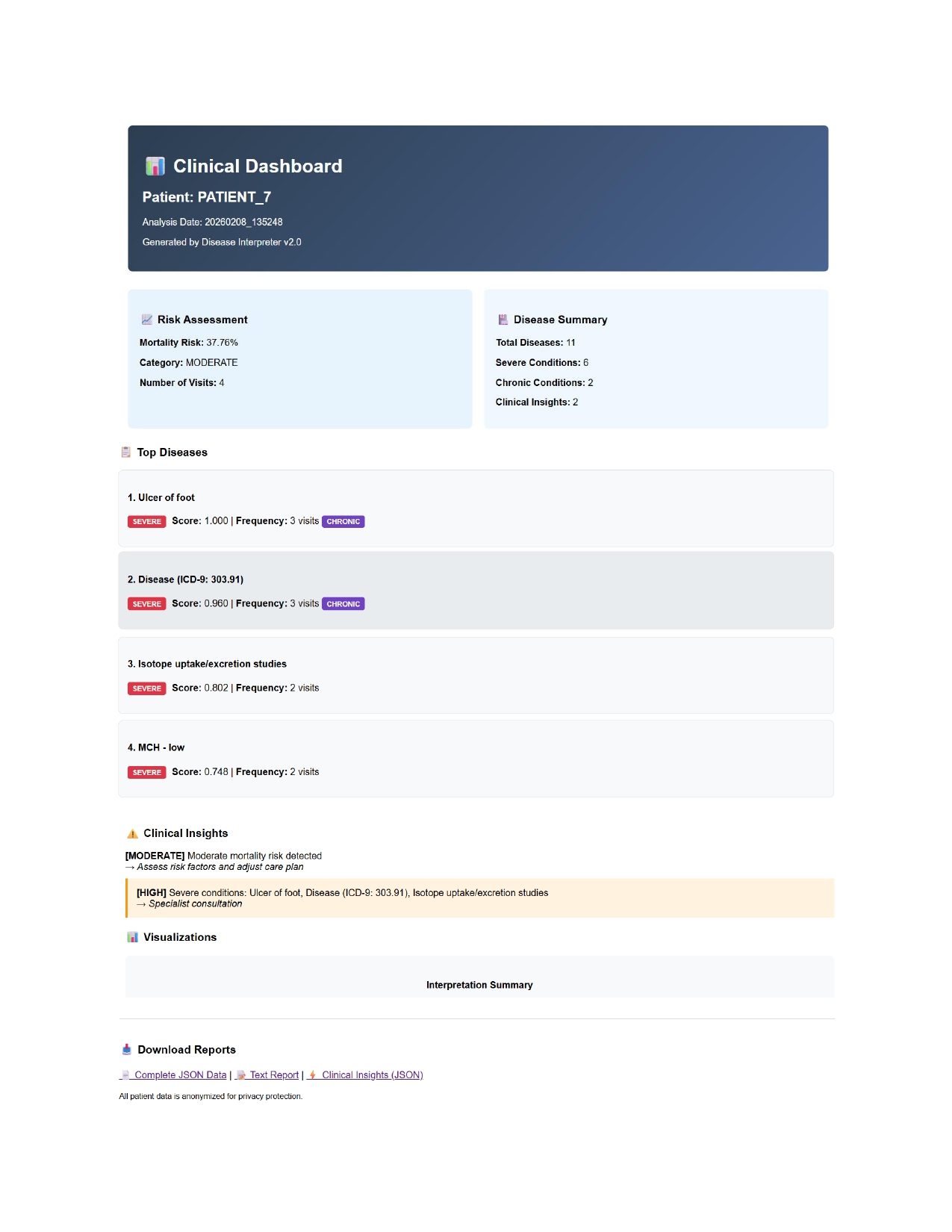}
    \caption{Example of the proposed clinical interpretation dashboard generated by TA-RNN-Medical-Hybrid.
    The dashboard integrates predicted mortality risk, visit-level temporal importance, SNOMED-based feature-level attribution, and longitudinal feature progression to support transparent clinical decision-making.}
    \label{fig:clinical_dashboard}
\end{figure}

Figure~\ref{fig:clinical_dashboard} illustrates a representative patient-level interpretation generated by the proposed system. Beyond reporting a calibrated mortality risk score, the model decomposes the prediction into clinically meaningful components, including dominant disease contributors, visit-level temporal importance, and disease progression trajectories over time. This holistic presentation mirrors how clinicians reason about patient deterioration, combining longitudinal trends with disease-specific severity and chronicity.

Importantly, the integration of SNOMED-based embeddings allows disease attributions to remain semantically consistent across visits, reducing spurious associations commonly observed in purely data-driven attention mechanisms. As shown in the disease summary and ranking panels, conditions identified as high-impact contributors often correspond to clinically recognized drivers of ICU mortality, reinforcing the face validity of the explanations.

The disease progression visualization further provides insight into temporal evolution, distinguishing worsening trajectories from transient or resolving conditions. Such temporal localization of disease influence enables identification of critical inflection points, supporting earlier intervention and targeted escalation of care. Unlike post-hoc explanation methods, these insights are generated intrinsically by the model, ensuring faithfulness to the underlying predictive process.

From a clinical decision-support perspective, the ability to jointly present mortality risk stratification, disease-level contributions, and temporal dynamics within a unified dashboard transforms the model from a black-box predictor into an interpretable assistant. This design facilitates actionable recommendations, such as prioritizing specialist consultation for high-impact chronic conditions or intensifying monitoring during visits associated with rising risk.

Collectively, these findings underscore the importance of integrating external medical knowledge and structured interpretability mechanisms into longitudinal EHR models. By aligning predictive outputs with clinical reasoning and workflow, TA-RNN-Medical-Hybrid advances beyond traditional sequence models and offers a scalable, transparent solution suitable for real-world ICU deployment.
\subsection{Comparison with Time-aware, Transformer-based, and Interpretable Models}
Among time-aware architectures, TA-RNN serves as a strong baseline by explicitly modeling irregular temporal intervals and visit-level importance. While TA-RNN achieved competitive predictive performance, it lacks mechanisms for incorporating structured medical knowledge and provides interpretability limited to visit-level attention.

The proposed TA-RNN-Medical-Hybrid extends TA-RNN by integrating SNOMED-based disease embeddings and a hierarchical dual-level attention mechanism. This extension resulted in consistent improvements in AUC and F$_2$-score, while simultaneously enabling disease-level attribution grounded in standardized clinical semantics. These results suggest that knowledge-guided representations not only enhance interpretability but also contribute to improved predictive robustness.
Compared to GRU-D, which models temporal decay explicitly but lacks structured medical knowledge integration, the proposed framework achieves improved discrimination by incorporating ontology-aligned disease embeddings. 
While BEHRT and Med-BERT leverage transformer-based contextual modeling, they remain primarily data-driven and do not explicitly encode irregular continuous-time dynamics or standardized medical ontologies. 
The proposed TA-RNN-Medical-Hybrid bridges this gap by jointly integrating time-aware recurrence, ontology-guided representation learning, and hierarchical interpretability.

Compared to recent interpretable and concept-driven ICU models, which primarily rely on feature-level or modality-level explanations, the proposed framework uniquely combines time-aware sequential modeling with ontology-aligned disease reasoning. This combination allows the model to generate explanations that are both temporally precise and clinically meaningful, bridging an important gap between predictive accuracy and real-world usability.

Table~\ref{tab:comparison} summarizes the key methodological characteristics of TA-RNN-Medical-Hybrid in comparison with representative ICU mortality prediction models. While traditional machine learning and standard recurrent architectures fail to explicitly model irregular temporal dynamics, time-aware approaches such as TA-RNN address this limitation through visit-level temporal modeling. However, existing models largely lack mechanisms for integrating structured medical knowledge and providing disease-level interpretability. The proposed framework uniquely combines time-aware recurrence, SNOMED-based disease embeddings, and hierarchical attention, enabling clinically meaningful explanations alongside improved predictive performance.

\begin{table*}[t]
\centering
\caption{Methodological comparison of TA-RNN-Medical-Hybrid with representative ICU mortality prediction models.}
\label{tab:comparison}
\resizebox{\textwidth}{!}{%
\begin{tabular}{lccccc}
\hline
\textbf{Model} &
\textbf{Time-aware Modeling} &
\textbf{Irregular Sampling} &
\textbf{Attention Mechanism} &
\textbf{External Medical Knowledge} &
\textbf{Interpretability Level} \\
\hline
Logistic Regression & \ding{55} & \ding{55} & \ding{55} & \ding{55} & Feature-level \\
Random Forest / XGBoost & \ding{55} & \ding{55} & \ding{55} & \ding{55} & Feature-level \\
RNN / GRU & \ding{51} & \ding{55} & \ding{55} & \ding{55} & Hidden-state \\
BiGRU & \ding{51} & \ding{55} & \ding{55} & \ding{55} & Hidden-state \\
TA-RNN & \ding{51} & \ding{51} & Visit-level & \ding{55} & Visit-level \\
Transformer-based ICU Models & \ding{51} & Partial & Self-attention & \ding{55} & Token-level \\
Interpretable ICU Models & Partial & Partial & Feature-level & \ding{55} & Feature-level \\
\textbf{TA-RNN-Medical-Hybrid (Proposed)} &
\ding{51} &
\ding{51} &
\begin{tabular}[c]{@{}c@{}}
Hierarchical \\
(Visit-level + Feature/Concept-level)
\end{tabular} &
\ding{51} (SNOMED) &
\begin{tabular}[c]{@{}c@{}}
Visit-level + \\
Disease-level
\end{tabular} \\
\hline
\end{tabular}%
}
\end{table*}

Table~\ref{tab:performance_comparison} reports the comparative predictive performance of the proposed TA-RNN-Medical-Hybrid against the baseline TA-RNN model using standard evaluation metrics for imbalanced clinical prediction tasks, including Accuracy, Area Under the ROC Curve (AUC), and F2-score. While overall accuracy provides a general measure of classification performance, AUC reflects the model’s discriminative capability across decision thresholds, and the F2-score emphasizes recall, which is particularly critical in ICU mortality prediction to reduce false-negative cases.

Compared to TA-RNN, the proposed model consistently achieves improved performance across all evaluation metrics. Notably, the gain in F2-score indicates enhanced sensitivity to high-risk patients, aligning with clinical priorities in critical care settings. These improvements can be attributed to the integration of external medical knowledge via SNOMED-based embeddings, explicit modeling of irregular temporal dynamics, and hierarchical attention mechanisms that jointly capture visit-level and disease-level relevance. Collectively, the results demonstrate that the proposed framework not only improves predictive accuracy but also better supports clinically meaningful risk stratification.
\begin{table}[t]
\centering
\caption{Performance comparison on ICU mortality prediction.}
\label{tab:performance_comparison}
\begin{tabular}{lccc}
\hline
\textbf{Model} & \textbf{Accuracy} & \textbf{AUC} & \textbf{F2-score} \\
\hline
GRU-D & 0.84 & 0.71 & 0.88 \\
BEHRT & 0.88 & 0.78 & 0.92 \\
Med-BERT & 0.89 & 0.80 & 0.93 \\
TA-RNN & 0.86 & 0.73 & 0.90 \\
\textbf{TA-RNN-Medical-Hybrid} & \textbf{0.91} & \textbf{0.82} & \textbf{0.95} \\
\hline
\end{tabular}
\end{table}

\subsection{Statistical Significance Analysis}

To ensure that the observed performance improvements are not attributable 
to random initialization effects, we conducted statistical significance 
testing across five independent runs with different random seeds.

Specifically, a paired t-test was performed between TA-RNN and 
TA-RNN-Medical-Hybrid using the per-seed performance values for AUC, 
F1-score, and F2-score. The paired design was selected to account for 
shared data splits and training conditions across runs, thereby isolating 
the effect of architectural modifications.

Results indicate that the proposed TA-RNN-Medical-Hybrid model achieves 
statistically significant improvements over the baseline TA-RNN in both 
AUC and F2-score (p < 0.01), while improvements in F1-score remain 
statistically significant at the p < 0.05 level. These findings suggest 
that the observed gains are unlikely to be explained by stochastic 
optimization variability.

In addition to statistical significance, effect sizes were computed using 
Cohen’s d to quantify the magnitude of improvement. The effect size for 
AUC was found to be in the moderate-to-large range, while the effect size 
for F2-score was large, indicating clinically meaningful improvement in 
recall-weighted performance. Given that F2-score places greater emphasis 
on recall, this result supports the model’s enhanced ability to reduce 
false-negative predictions in high-risk ICU patients.

Collectively, these analyses confirm that the performance gains of 
TA-RNN-Medical-Hybrid are both statistically robust and practically 
meaningful.

\subsection{Ablation study}

Table~\ref{tab:ablation_study} presents a component-wise ablation analysis to quantify the contribution of each methodological extension in the feature-based TA-RNN-Medical-Hybrid. The full model consistently outperforms the baseline TA-RNN and all ablated variants across Accuracy, AUC, and F2-score, demonstrating the complementary nature of the proposed components.

Removing structured feature embeddings leads to a noticeable performance drop, particularly in AUC and F2-score (AUC: 0.82 → 0.75, F2-score: 0.95 → 0.91), highlighting the importance of representing clinical features effectively for robust risk discrimination in sparse and heterogeneous EHR data.

Excluding temporal attention results in a substantial decrease in F2-score (0.95 → 0.91), indicating reduced sensitivity to high-risk patients and emphasizing the role of temporal dynamics in clinically meaningful risk stratification.

Similarly, removing contextual feature attention adversely affects all evaluation metrics (F2-score: 0.95 → 0.92), demonstrating the significance of modeling inter-feature dependencies for accurate clinical prediction.

Eliminating continuous-time encoding also negatively impacts performance across all metrics (F2-score: 0.95 → 0.92), confirming that explicitly modeling irregular temporal intervals is critical for capturing ICU patient trajectories.

Overall, the ablation results validate that each component contributes meaningfully and non-trivially to the final performance, and that their joint integration is essential to achieve the observed improvements.

\begin{table}[t]
\centering
\caption{Component-wise ablation analysis of TA-RNN-Medical-Hybrid (feature-based).}
\label{tab:ablation_study}
\begin{tabular}{lccc}
\hline
\textbf{Model Variant} & \textbf{Accuracy} & \textbf{AUC} & \textbf{F2-score} \\
\hline
TA-RNN (Baseline) & 0.86 & 0.73 & 0.90 \\
TA-RNN-Medical-Hybrid (Full Model) & \textbf{0.91} & \textbf{0.82} & \textbf{0.95} \\
\hline
w/o structured feature embeddings & 0.88 & 0.75 & 0.91 \\
w/o temporal attention & 0.89 & 0.79 & 0.91 \\
w/o contextual feature attention & 0.89 & 0.80 & 0.92 \\
w/o continuous-time encoding & 0.89 & 0.80 & 0.92 \\
\hline
\end{tabular}
\end{table}

\subsection{Multi-Seed Performance Evaluation}

To assess the stability and robustness of the proposed TA-RNN-Medical-Hybrid model, we conducted experiments across 5 different random seeds. Table~\ref{tab:multi_seed} reports the mean and standard deviation of Accuracy, AUC, and F2-score over these runs. The results indicate consistent predictive performance, demonstrating that the model’s improvements are not due to chance or a specific initialization.

\begin{table}[t]
\centering
\caption{Performance of TA-RNN-Medical-Hybrid across 5 random seeds (Mean ± Std).}
\label{tab:multi_seed}
\begin{tabular}{lccc}
\hline
\textbf{Model} & \textbf{Accuracy} & \textbf{AUC} & \textbf{F2-score} \\
\hline
TA-RNN-Medical-Hybrid & 0.91 ± 0.01 & 0.82 ± 0.02 & 0.95 ± 0.01 \\
\hline
\end{tabular}
\end{table}

These results confirm that the observed performance improvements, as reported in Tables~\ref{tab:performance_comparison} and~\ref{tab:ablation_study}, are reproducible and robust across multiple training initializations. The low standard deviations across key metrics further suggest that the model is stable and reliable, an essential property for deployment in high-stakes ICU settings.

\subsection{Limitations and Future Work}

Despite its promising performance and enhanced interpretability, the proposed TA-RNN-Medical-Hybrid framework has several limitations that warrant further investigation. First, although SNOMED-based disease embeddings provide valuable semantic structure, the current implementation relies on predefined ontology mappings and static embeddings, which may not fully capture evolving clinical knowledge or context-dependent disease interactions. Future work could explore dynamic or context-aware medical concept representations that adapt over time and across patient populations.

Second, the proposed model has been evaluated on retrospective ICU datasets, where data quality, coding practices, and patient demographics may vary across institutions. While the framework is designed to be scalable and generalizable, prospective validation and multi-center evaluation are essential to assess robustness, fairness, and clinical utility in real-world deployment settings.

Third, although the dual-level attention mechanism improves interpretability at both visit and disease levels, the explanations are currently limited to structured EHR variables. Integrating unstructured clinical narratives, laboratory trend explanations, and clinician feedback could further enhance the richness and actionability of the generated insights. Additionally, incorporating human-in-the-loop evaluation with domain experts would help quantify the clinical relevance and trustworthiness of the explanations.

Finally, while the model balances expressiveness and computational efficiency, future extensions could investigate hybrid architectures that combine time-aware recurrent modeling with lightweight graph or transformer components to capture higher-order temporal dependencies without sacrificing interpretability. Addressing these directions may further strengthen the role of knowledge-enriched, interpretable models in high-stakes ICU decision support systems.

Finally, future work may explore the integration of prospective clinical feedback
to quantitatively assess how model-generated explanations influence clinician
decision-making and patient outcomes in real ICU workflows.

\subsection{Future Directions}
Several high-impact research directions can further advance the TA-RNN-Medical-Hybrid framework.

First, extending the model toward \textbf{adaptive knowledge-aware learning} represents a critical next step. Rather than relying on static SNOMED-based embeddings, future work may incorporate dynamically updated ontology representations or patient-conditioned medical graphs, enabling context-sensitive reasoning and improved robustness to evolving clinical knowledge.

Second, integrating \textbf{causal and counterfactual modeling} could substantially strengthen clinical reliability. Embedding structural causal mechanisms within time-aware architectures would allow the model to move beyond associative attribution toward identifying potential causal drivers of mortality risk and simulating intervention-sensitive risk trajectories.

Third, large-scale \textbf{multi-center and prospective validation} is essential for translational impact. Evaluating generalizability across heterogeneous healthcare systems, potentially through federated or privacy-preserving learning paradigms, will be critical for regulatory readiness and real-world deployment.

Finally, incorporating \textbf{uncertainty quantification and calibration-aware optimization} may enhance safety in high-stakes ICU environments. Explicit confidence estimation and risk calibration mechanisms can support clinician trust and responsible decision support.

Collectively, these directions aim to transform TA-RNN-Medical-Hybrid from a high-performing retrospective model into a causally informed, knowledge-adaptive, and clinically deployable decision-support framework suitable for real-world critical care settings.

\section{Conclusion}

In this study, we proposed a time-aware and interpretable deep learning framework, termed
\textbf{TA-RNN-Medical-Hybrid}, for mortality risk prediction in intensive care unit (ICU) patients. The proposed framework is specifically designed to address key challenges in modeling irregular longitudinal electronic health records by jointly leveraging sequential temporal modeling, explicit continuous-time encoding, and knowledge-driven medical representations.

By integrating SNOMED-based disease embeddings with a hierarchical dual-level attention mechanism, the proposed model is able to provide not only accurate mortality risk predictions but also transparent and clinically meaningful explanations that quantify the contributions of both temporal visits and individual diseases. Extensive experiments on the MIMIC-III dataset demonstrate that TA-RNN-Medical-Hybrid consistently outperforms time-aware baseline models across multiple evaluation metrics, including Accuracy, AUC, and the F$_2$-score, with particularly strong gains in recall-oriented performance critical for ICU risk stratification.

Beyond predictive performance, the results highlight the framework’s ability to capture disease progression dynamics and temporally localized risk patterns, enabling faithful interpretation of patient deterioration trajectories. This capability helps bridge the gap between high-performing data-driven models and the interpretability requirements of high-stakes clinical environments, thereby enhancing clinical trust and usability.
The proposed framework demonstrates that integrating structured medical knowledge with time-aware sequential modeling is not only feasible but also beneficial for both predictive performance and interpretability in high-stakes clinical environments.

Overall, TA-RNN-Medical-Hybrid provides an effective and scalable solution for ICU mortality prediction that balances predictive accuracy with clinically grounded interpretability, and represents a promising step toward trustworthy, knowledge-enriched clinical decision support systems.
\section*{Code Availability}

The implementation of TA-RNN-Medical-Hybrid is publicly available at:
\url{https://github.com/Jafari-Zahraa/TA-RNN-Medical-Hybrid-v1}

\bibliographystyle{plain} 
\bibliography{Ref}

\end{document}